# Byzantine Stochastic Gradient Descent

Dan Alistarh
dan.alistarh@ist.ac.at
IST Austria

Zeyuan Allen-Zhu
zeyuan@csail.mit.edu
Microsoft Research AI

Jerry Li*
jerryzli@mit.edu
MIT CSAIL

March 23, 2018

## Abstract

This paper studies the problem of distributed stochastic optimization in an adversarial setting where, out of the $m$ machines which allegedly compute stochastic gradients every iteration, an $\alpha$-fraction are Byzantine, and can behave arbitrarily and adversarially. Our main result is a variant of stochastic gradient descent (SGD) which finds $\varepsilon$-approximate minimizers of convex functions in $T = \widetilde{O}\big(\frac{1}{\varepsilon^2 m} + \frac{\alpha^2}{\varepsilon^2}\big)$ iterations. In contrast, traditional mini-batch SGD needs $T = O\big(\frac{1}{\varepsilon^2 m}\big)$ iterations, but cannot tolerate Byzantine failures. Further, we provide a lower bound showing that, up to logarithmic factors, our algorithm is information-theoretically optimal both in terms of sampling complexity and time complexity.

## 1 Introduction

Data used in machine learning applications is becoming increasingly decentralized. This can be either because the data is naturally distributed—for instance, in applications such as *federated learning*, where data collection is crowdsourced [20]—or because data is partitioned across machines in order to parallelize computation, e.g. [4]. In both these cases, the basic setting is the same: data is spread across a large number of worker machines, with the goal of learn some global function, defined over all of the data points.

Fault-tolerance is a critical concern in such distributed settings. For instance, machines in a data center may crash, or fail in unpredictable ways. Even worse, when crowdsourcing data collection, one can no longer fully trust the data, and must be able to tolerate a fraction of *adversarial* workers, sending corrupted or even malicious data. This *Byzantine* failure model—where a small fraction of bad workers are allowed to behave arbitrarily—has a rich history in the distributed computing literature [22]. By contrast, the design of machine learning algorithms which are robust to such Byzantine failures is a relatively recent topic, but is rapidly becoming a major research direction at the intersection of machine learning, distributed computing, and security.

Algorithms in this setting should be measured against two fundamental complexity criteria:

- SAMPLE COMPLEXITY: algorithms should achieve high test accuracy with few data samples, even when a relatively large fraction of the workers are Byzantine.

- SCALABILITY AND COMPUTATIONAL COMPLEXITY: Algorithms should preserve (as much as possible) the runtime speedup achieved by distributing computation across multiple workers.

---

*Supported by NSF CAREER Award CCF-1453261, CCF-1565235, a Google Faculty Research Award, and an NSF Graduate Research Fellowship.

Ideally, we would like to achieve algorithms which, in the presence of Byzantine failures, (1) still require an information-theoretically optimal number of samples per machine to achieve a target error, and (2) run in a constant number of passes over the data. In addition, as the fraction of Byzantine machines tends to zero, we wish to match the sample complexity and time complexity of the best (non-Byzantine) distributed algorithms. In a nutshell, we want the overhead of tolerating Byzantine faults to be limited, and to be negligible if all nodes are non-faulty.

Another important consideration in the design of these algorithms is that they should remain useful in high dimensions. Since nowadays most datasets are high dimensional, it is critical to have algorithms with good complexity in such settings. In particular, having sample and/or time complexities which are off from the (information-theoretically) optimal by factors polynomial in the dimension can significantly decrease the practicality of such algorithms.

## 1.1 The Model

In this paper, we study stochastic optimization in a Byzantine setting. There is an unknown distribution $\mathcal{D}$ over functions $\mathbb{R}^d \to \mathbb{R}$, and we wish to minimize $f(x) \stackrel{\text{def}}{=} \mathbb{E}_{s \sim \mathcal{D}}[f_s(x)]$.

We assume a standard setting with $m$ worker machines and a master (coordinator) machine, and that an $\alpha$-fraction of the workers may be Byzantine (where $\alpha < 1/2$). Each worker has access to $T$ sample functions from the distribution $\mathcal{D}$. We proceed in iterations, structured as follows: in each iteration, workers perform some local computation, then synchronously send information to the master, which compiles the information and sends new information to the workers. At the end, the master should output an approximate minimizer of the function $f$.

While our lower bounds (i.e., negative results) will apply for this extremely general setting, our upper bounds (i.e., algorithms) will be expressed in the standard framework of distributed stochastic gradient methods. Specifically, in each iteration $k$ of the algorithm, the master broadcasts the current iterate $x_k \in \mathbb{R}^d$ to worker machines, and each worker is supposed to compute a stochastic gradient at $x_k$ and return it to the master:

- a good (i.e., non-Byzantine) worker machine returns $\nabla f_s(x_k)$ for a random sample $s \sim \mathcal{D}$, but
- a Byzantine worker machine may adversarially return any vector.

This stochastic optimization framework is general and very well studied, and captures many important problems such as regression, learning SVMs, logistic regression, and training deep neural networks. Traditional methods such as mini-batch *stochastic gradient descent (SGD)* naturally work well in distributed settings, however, are easily seen to be vulnerable to even a single Byzantine failure.

We now translate the aforementioned two abstract complexity criteria into this setting. *Sample complexity* is measured as the number of functions $f_s(\cdot)$ we accessed. Since every machine gets one sample per iteration, minimizing sample complexity is equivalent to minimizing the number of iterations. The *time complexity* of the algorithm is determined by the number of iterations required, as well as the per-iteration computation time. Since each iteration usually requires only one stochastic gradient computation (and other manipulations of the same order of complexity), minimizing running time is also equivalent to minimizing the number of iterations.

There are also other important considerations in the design of distributed algorithms. Notably, per-iteration communication cost is often a bottleneck to scalability, and minimizing it is quite important. Our algorithms will require the workers to only send a single gradient every iteration, and therefore will automatically have low per-iteration communication requirements.

## 1.2 Our Results

In this work, we study the convex formulation of this Byzantine stochastic optimization: we assume $f(x)$ is convex but each $f_s(x)$ may not necessarily be convex. We provide the first algorithms that, in the presence of Byzantine machines (and up to log factors and lower-order terms),

(1) achieve the optimal sample complexity,

(2) achieve the optimal number of stochastic gradient computations,

(3) match the sample and time complexity of traditional SGD as $\alpha \to 0$, and

(4) achieve (1)-(3) even as the dimension grows, without losing any additional dimension factors.

In addition, our algorithms are optimally-robust, supporting a fraction of $\alpha < 1/2$ Byzantine workers. Despite significant recent interest, e.g. [8, 11, 15, 33, 34, 38, 39], to the best of our knowledge, prior to our work there were no algorithms for stochastic optimization in high dimensions that achieved any of the four objectives highlighted above. Previous algorithms either provided weak robustness guarantees, or had sample or time complexities which degrade polynomially with the dimension $d$ or with the error $\varepsilon$.

**Theorem Statements.** We show provable guarantees for a number of interesting settings, in particular, for both smooth and non-smooth functions, and for convex and strongly convex functions. We summarize our results in the smooth setting as follows:[1]

> **Theorem 1** (informal). *If $f(x)$ is smooth, convex and $\alpha$-fraction of the machines are Byzantine for $\alpha < 1/2$, then* `ByzantineSGD` *finds a point $x$ with $f(x) - f(x^*) \le \varepsilon$ in $T$ iterations where*
> $$T = \widetilde{O}\Big(\frac{1}{\varepsilon} + \frac{1}{\varepsilon^2 m} + \frac{\alpha^2}{\varepsilon^2}\Big) \quad \text{or} \quad T = \widetilde{O}\Big(\frac{1}{\sigma} + \frac{1}{\sigma \varepsilon m} + \frac{\alpha^2}{\sigma \varepsilon}\Big) \text{ if } f(x) \text{ is } \sigma\text{-strongly convex.}$$

One should compare Theorem 1 to the traditional mini-batch version of SGD (that works only for $\alpha = 0$, the non-Byzantine setting), where the number of iterations is

$$T = O\Big(\frac{1}{\varepsilon} + \frac{1}{\varepsilon^2 m}\Big) \quad \text{or} \quad T = \widetilde{O}\Big(\frac{1}{\sigma} + \frac{1}{\sigma \varepsilon m}\Big) \text{ if } f(x) \text{ is } \sigma\text{-strongly convex.}$$

In other words, the additional loss incurred by Byzantine machines is the additive term $\frac{\alpha^2}{\varepsilon^2}$ or $\frac{\alpha^2}{\sigma \varepsilon}$. We also show a matching information-theoretic lower bound for this term:

> **Theorem 2** (informal). *In the presence of $\alpha$-fraction of Byzantine machines, if an algorithm finds a point $x$ with $f(x) - f(x^*) \le \varepsilon$ by acquiring $T$ samples from $\mathcal{D}$ per machine, then*
> $$T = \Omega\Big(\frac{1}{\varepsilon^2 m} + \frac{\alpha^2}{\varepsilon^2}\Big) \quad \text{or} \quad T = \Omega\Big(\frac{1}{\sigma \varepsilon m} + \frac{\alpha^2}{\sigma \varepsilon}\Big) \text{ if } f(x) \text{ is } \sigma\text{-strongly convex.}$$

Intuitively, the lower bound term $\frac{1}{\varepsilon^2 m}$ (resp. $\frac{1}{\sigma \varepsilon m}$) is due to the folklore sampling complexity result of SGD (in the non-Byzantine setting). Namely, at least $\frac{1}{\varepsilon^2}$ (resp. $\frac{1}{\sigma \varepsilon}$) samples of the functions must be needed in order to find an $\varepsilon$-approximate minimizer of $f(x)$. (See for instance Woodworth and Srebro [37].) Our main contribution in Theorem 2 is the additional term $\frac{\alpha^2}{\varepsilon^2}$ (resp. $\frac{\alpha^2}{\sigma \varepsilon}$) which shows that the error introduced by `ByzantineSGD` is *necessary*.

*Remark* 1.1. While the second and third terms in Theorem 1 are optimal owing to Theorem 2, the first term, namely $\frac{1}{\varepsilon}$ or $\frac{1}{\sigma}$, is not. (It is usually not the dominant term anyways.) One can use acceleration methods [26] to improve it to $\frac{1}{\sqrt{\varepsilon}}$ or $\frac{1}{\sqrt{\sigma}}$ (cf. [16]), and this is optimal (in terms of number of stochastic gradient computations as opposed to number of sampled functions) when the dimension is sufficiently high [26].

---

[1] The non-smooth statements are similar and can be found in later sections. In this informal statement, following tradition in optimization literatures, we have hidden the dependency on the smoothness, the diameter, and the variance parameters. See later sections for the full specification of such parameters. We use the $\widetilde{O}$ notation to hide a logarithmic factor in the parameters.

*Remark* 1.2. Parallel speedup is usually captured by how $T$ (usually known as parallel running time) decreases as a function of $m$. In the non-Byzantine setting, (mini-batch) SGD satisfies the property that $T$ improves by a factor of $\Omega(m)$ as long as $m \leq \frac{1}{\varepsilon}$. This is one of the reasons SGD is favored in large-scale parallel optimization. In the Byzantine setting, our `ByzantineSGD` satisfies the property that $T$ improves by a factor of $\Omega(m)$ as long as $m \leq \min\{\frac{1}{\varepsilon}, \frac{1}{\alpha^2}\}$. In other words, we have preserved (as much as we can) the runtime speedup achieved by traditional methods in the non-Byzantine setting.

## 1.3 Our Techniques and Algorithms

One naive way to deal with Byzantine workers is to perform a robust aggregation step to compute gradients, such as median of means. Namely, for each (good) worker machine $i \in [m]$, whenever a query point $x_k$ is provided by the master, the worker takes $n$ stochastic gradient samples and computes their average, which we call $v_i$. If $n = \widetilde{\Theta}(\varepsilon^{-2})$, one can readily show that for each good machine $i$, it holds that $\|v_i - \nabla f(x_k)\| \leq \varepsilon$ with high probability. Therefore, in each iteration $k$, we can determine a vector $v_{\mathsf{med}} \in \{v_1, \ldots, v_m\}$ satisfying $\|v_{\mathsf{med}} - \nabla f(x_k)\| \leq 2\varepsilon$, and move in the negative direction of $v_{\mathsf{med}}$.

Unfortunately, the above idea requires too many computations of stochastic gradients. For example, in the non-strongly convex setting, each worker machine needs to compute $\varepsilon^{-2}$ stochastic gradients per iteration, and the overall number of iterations will be at least $\varepsilon^{-1}$.[2] This amounts to a total of at least $\varepsilon^{-3}$ in the total complexity.

We take a different path in this paper. We run the algorithm for $T$ iterations, where each machine $i \in [m]$ only computes *one* stochastic gradient per iteration. Let $v_i^{(k)}$ be the stochastic gradient allegedly computed by machine $i \in [m]$ at iteration $k \in [T]$. By martingale concentration, $B_i \stackrel{\text{def}}{=} \frac{v_i^1 + \cdots + v_i^{(T)}}{T}$ should concentrate around $B_\star \stackrel{\text{def}}{=} \frac{\nabla f(x_1) + \cdots + \nabla f(x_T)}{T}$ for each good machine $i$, up to an additive error $\frac{1}{\sqrt{T}}$. In other words, if $\|B_i - B_\star\| > \frac{1}{\sqrt{T}}$ for some machine $i$, we can safely declare $i$ is a Byzantine machine. Two technical issues remain:

- If we discover some machine is Byzantine, what should we do? If we restart the algorithm using the remaining machines, the total complexity blows up.

  In our algorithm `ByzantineSGD`, we compute $B_i^{(k)} \stackrel{\text{def}}{=} \frac{v_i^1 + \cdots + v_i^{(k)}}{k}$ on the fly as $k$ increases. If at any iteration $k$ we find out $B_i^{(k)}$ is too far away from $B_\star^{(k)}$, we declare $i$ is a Byzantine machine and remove it from *future consideration*. We show that restarting is not necessary.

- Some Byzantine machine $i$ may hide itself without violating the criterion "$\|B_i - B_\star\| \leq \frac{1}{\sqrt{T}}$."

  In our algorithm `ByzantineSGD`, in addition to $B_i^{(k)}$, we construct $A_i^{(k)} \stackrel{\text{def}}{=} \frac{\langle v_i^{(1)}, x_1 - x_0 \rangle + \cdots + \langle v_i^{(k)}, x_k - x_0 \rangle}{k}$ and check whether "$|A_i^{(k)} - A_\star^{(k)}| > \frac{1}{\sqrt{k}}$." It turns out any good machine always satisfy the both criteria. More importantly, even though there may still be Byzantine machines which satisfy both criteria, their contributions to the loss in accuracy can be proven negligible.

---

[2]This is because, even when $f_s(x) = f(x)$ (so there is no stochasticity) and when $\alpha = 0$ (so there is no Byzantine machine), gradient descent converges in $\varepsilon^{-1}$ iterations.

| algorithm | # sampled functions per machine | total work | per-iteration per-machine work |
|---|---|---|---|
| SGD ($\alpha = 0$ only) (folklore) | $O\bigl(\frac{1}{\varepsilon} + \frac{1}{\varepsilon^2 m}\bigr)$ | $O\bigl(\frac{m}{\varepsilon} + \frac{1}{\varepsilon^2}\bigr)$ | 1 |
| `ByzantineSGD` (Theorem 3.8) | $\widetilde{O}\bigl(\frac{1}{\varepsilon} + \frac{1}{\varepsilon^2 m} + \frac{\alpha^2}{\varepsilon^2}\bigr)$ | $\widetilde{O}\bigl(\frac{m}{\varepsilon} + \frac{1}{\varepsilon^2} + \frac{\alpha^2 m}{\varepsilon^2}\bigr)$ | 1 |
| GD ($\alpha = 0$ only) (folklore) | $\widetilde{O}\bigl(1 + \frac{1}{\varepsilon^2 m}\bigr)$ | $\widetilde{O}\bigl(\frac{m}{\varepsilon} + \frac{1}{\varepsilon^3}\bigr)$ | $1 + \widetilde{O}\bigl(\frac{1}{\varepsilon^2 m}\bigr)$ |
| Median-GD (Yin et al. [39]) | $\widetilde{O}\bigl(1 + \frac{d}{\varepsilon^2 m} + \frac{\alpha^2}{\varepsilon^2}\bigr)$ | $\widetilde{O}\bigl(\frac{m}{\varepsilon} + \frac{d}{\varepsilon^3} + \frac{\alpha^2 m}{\varepsilon^3}\bigr)$ | $1 + \widetilde{O}\bigl(\frac{d}{\varepsilon^2 m} + \frac{\alpha^2}{\varepsilon^2}\bigr)$ |
| folklore (c.f. [37, Theorem 11]) | $\Omega\bigl(\frac{1}{\varepsilon^2 m}\bigr)$ | $\Omega\bigl(\frac{1}{\varepsilon^2}\bigr)$ | |
| this paper (Theorem 5.4) | $\Omega\bigl(\frac{1}{\varepsilon^2 m} + \frac{\alpha^2}{\varepsilon^2}\bigr)$ | $\Omega\bigl(\frac{1}{\varepsilon^2} + \frac{\alpha^2 m}{\varepsilon^2}\bigr)$ | |
| ⇈ convex ⇈ | | ⇊ $\sigma$-strongly convex ⇊ | |
| SGD ($\alpha = 0$ only) (folklore) | $O\bigl(\frac{1}{\sigma} + \frac{1}{\sigma\varepsilon m}\bigr)$ | $O\bigl(\frac{m}{\sigma} + \frac{1}{\sigma\varepsilon}\bigr)$ | 1 |
| `ByzantineSGD` (Theorem 4.2) | $\widetilde{O}\bigl(\frac{1}{\sigma} + \frac{1}{\sigma\varepsilon m} + \frac{\alpha^2}{\sigma\varepsilon}\bigr)$ | $\widetilde{O}\bigl(\frac{m}{\sigma} + \frac{1}{\sigma\varepsilon} + \frac{\alpha^2 m}{\sigma\varepsilon}\bigr)$ | 1 |
| GD ($\alpha = 0$ only) (folklore) | $O\bigl(1 + \frac{1}{\sigma\varepsilon m}\bigr)$ | $O\bigl(\frac{m}{\sigma} + \frac{1}{\sigma^2\varepsilon}\bigr)$ | $1 + O\bigl(\frac{1}{\sigma\varepsilon m}\bigr)$ |
| Median-GD (Yin et al. [39]) | $\widetilde{O}\bigl(1 + \frac{d}{\sigma\varepsilon m} + \frac{\alpha^2}{\sigma\varepsilon}\bigr)$ | $\widetilde{O}\bigl(\frac{m}{\sigma} + \frac{d}{\sigma^2\varepsilon} + \frac{\alpha^2 m}{\sigma^2\varepsilon}\bigr)$ | $1 + \widetilde{O}\bigl(\frac{d}{\sigma\varepsilon m} + \frac{\alpha^2}{\sigma\varepsilon}\bigr)$ |
| folklore (c.f. [37, Appendix C.5]) | $\Omega\bigl(\frac{1}{\sigma\varepsilon m}\bigr)$ | $\Omega\bigl(\frac{1}{\sigma\varepsilon}\bigr)$ | |
| this paper (Theorem 5.5) | $\Omega\bigl(\frac{1}{\sigma\varepsilon m} + \frac{\alpha^2}{\sigma\varepsilon}\bigr)$ | $\Omega\bigl(\frac{1}{\sigma\varepsilon} + \frac{\alpha^2 m}{\sigma\varepsilon}\bigr)$ | |

Table 1: Comparison of Byzantine optimization for minimizing smooth convex functions $f(x) = \mathbb{E}_{s \sim \mathcal{D}}[f_s(x)]$.
  **Remark 1.** In this table, we have assumed parameters $L$ (smoothness), $\mathcal{V}$ (variance), and $D$ (diameter) as constants. The goal is to achieve $f(x) - f(x^*) \leq \varepsilon$, and $\sigma$ is the strong convexity parameter of $f(x)$.
  **Remark 2.** "# sampled functions" is the number of $f_s(\cdot)$ to sample for each machine.
  **Remark 3.** "total/per-iteration work" is in terms of the number of stochastic gradient computations $\nabla f_s(\cdot)$.
________________________________________________________________________

## 1.4 Related Work

**Byzantine GD.** The closest literature is the concurrent and independent work of Yin et al. [39]. They consider a similar Byzantine model, but for *gradient descent (GD)*. Below, we highlight the main distinctions.

MODEL DIFFERENCES. In their algorithm, *each* of the $m$ machines receives $n$ samples of functions upfront. In an iteration $k$, machine $i$ allegedly computes $n$ stochastic gradients at point $x_k$ and averages them (the $n$ stochastic gradient are with respect to the $n$ sampled functions stored on machine $i$). Then, their proposed algorithm aggregates all the average vectors from the $m$ machines, and performs a coordinate-wise median operation to determine the descent direction.

Their algorithm is a *Byzantine variant of GD*: a total of $nm$ functions are sampled upfront, and a total of $Tnm$ stochastic gradient computations (or equivalently $T$ full gradient computations) are performed. In contrast, our algorithm is a *Byzantine variant of SGD*: a total of $Tm$ functions are sampled and a total of $Tm$ stochastic gradient computations are performed.

TECHNICAL DIFFERENCES. To be robust against Byzantine machines, they average stochastic gradients within a single iteration and compare them across machines. In contrast, we average stochastic gradients (and other quantities) *across iterations*.

COMPLEXITY DIFFERENCES. In terms of sampling complexity (i.e., the number of functions $f_s(\cdot)$ to be sampled), their algorithm's complexity is higher by a linear factor in the dimension $d$ (see Table 1). This is in large part due to their coordinate-wise median operation. In high dimensions, this leads to sub-optimal statistical rates. In terms of total computational complexity, each iteration of theirs requires a full pass to the (sampled) dataset. In contrast, the entire run of `ByzantineSGD` requires only one pass.

ASSUMPTION DIFFERENCES. Their algorithm works under a weaker set of assumptions than ours. They assumed the stochastic error in gradients (namely, $\nabla f_s(x) - \nabla f(x)$) have bounded variance and skewness; in contrast, we simply assumed $\nabla f_s(x) - \nabla f(x)$ is bounded with probability 1. Our stronger assumption (which is anyways standard in certain literatures[3]) turns out to simplify our algorithm and analysis. We leave it as future work to extend `ByzantineSGD` to bounded skewness.

LOWER BOUND. Yin et al. [39] also provided a lower bound in terms of sampling complexity — the number of functions $f_s(\cdot)$ needed to be sampled in the presence of Byzantine machines. When translated to our language, the result is essentially the same as the strongly convex part of Theorem 2. (They do not provide lower bounds for non-strongly convex functions.)

**Byzantine Stochastic Optimization.** Recently there has been a lot of work on Byzantine stochastic optimization, and in particular, SGD [8, 11, 15, 33, 34, 38]. One of the first references to consider this setting is [15], which investigated distributed PCA and regression in the Byzantine distributed model. Their general framework has each machine running a robust learning algorithm locally, and aggregating results via a robust estimator. However, the algorithm requires careful parametrization of the sample size at each machine to obtain good error bounds, which renders it suboptimal with respect to sample complexity. Our work introduces new techniques which address both these limitations. References [33, 34] consider a similar setting. However, [33] focuses on the single-dimensional ($d = 1$) case. In [34], the authors consider the multi-dimensional setting, but only allow for optimizing over a bounded set of state candidates.

Reference [8] proposes a general Byzantine-resilient gradient aggregation rule called *Krum* for selecting a valid gradient update. This rule has local complexity $O(m^2(d + \log m))$, which makes it relatively expensive to compute when the $d$ and/or $m$ are large. Moreover, in each iteration the algorithm chooses a gradient corresponding to a constant number of correct workers, so it the scheme may not achieve speedup with respect to the number of distributed workers. Follow-up research [38] considered gradient aggregation rules in a *generalized* Byzantine setting where a subset of the messages sent between servers can be corrupted. The complexity of their selection rule can be as low as $\widetilde{O}(dm)$, but their approach is far from being sample-optimal. Recent work [11] leveraged the geometric median of means idea in a novel way, which allows it to be significantly more sample-efficient, and applicable for a wider range of parameters. At the same time, their technique only applies in the strongly convex setting, and is suboptimal in terms of convergence rate by a factor of $\sqrt{\alpha m}$.

Optimization and learning in the presence of adversarial noise is a well-studied problem, and in its most basic form dates back to foundational work of Huber and Tukey in the 60's and 70's [18, 36]. However, traditional efficient algorithms for robust optimization only worked in very limited models of corruption, or in very low dimensions [6, 7, 25, 27]. More recently, efficient algorithms for high dimensional optimization which are tolerant to a small fraction of adversarial corruptions have been developed [3, 10, 14, 19, 29], building on new algorithms for high dimensional robust statistics [7, 10, 12, 21]. This setting is quite similar to ours, but with some notable differences. In particular, in this setting, there are fundamental statistical barriers to the error any algorithm can achieve, no matter how many samples are taken.

**Distributed SGD.** There is an extremely rich literature studying the scalability and communication requirements of distributed SGD, e.g. [1, 9, 23, 24, 31, 32, 35, 40], in settings where all the workers are correct. Since none of these algorithms are designed to tolerate Byzantine failures, a survey of this area is beyond the scope of our work. However, it is an interesting question whether these communication-reduction techniques can be made robust to adversarial interference.

---
[3]For instance, some stochastic optimization literature assumes $f_s(x)$ is Lipschitz continuous, which implies $\nabla f_s(x) - \nabla f(x)$ is bounded.

**Approximation Criteria.** There are various notions for what "$\varepsilon$-approximation" mean for optimizing $f(x)$. In this paper, we have adopted the popular notion of finding a point $x$ with $f(x) - f(x^*) \leq \varepsilon$ (or in the strongly convex setting it is interchangeable with the criterion of $\frac{\sigma}{2}\|x - x^*\|^2 \leq \varepsilon$). In some applications, one may also consider finding a point $x$ with $\|\nabla f(x)\| \leq \varepsilon$. This task is harder and requires additional work on top of SGD [2] (even without Byzantine machines). Our results of this paper generalize to that setting, but for simplicity of exposition, we refrain from formally proving this.

## 1.5 Roadmap

We introduce notations and our formal Byzantine model for distributed stochastic optimization in Section 2. We analyze our `ByzantineSGD` algorithm in the non-strongly convex case in Section 3, and in the strongly convex case in Section 4. We prove lower bounds in Section 5.

## 2 Preliminaries

Throughout this paper, we denote by $\|\cdot\|$ the Euclidean norm and $[n] \stackrel{\text{def}}{=} \{1, 2, \ldots, n\}$. Recall some definitions on strong convexity, smoothness, and Lipschitz continuity (and they have other equivalent definitions, see textbook [26]).

**Definition 2.1.** *For a differentiable function $f \colon \mathbb{R}^d \to \mathbb{R}$,*
- *$f$ is $\sigma$-strongly convex if $\forall x, y \in \mathbb{R}^d$, it satisfies $f(y) \geq f(x) + \langle \nabla f(x), y - x \rangle + \frac{\sigma}{2}\|x - y\|^2$.*
- *$f$ is $L$-Lipschitz smooth (or $L$-smooth for short) if $\forall x, y \in \mathbb{R}^d$, $\|\nabla f(x) - \nabla f(y)\| \leq L\|x - y\|$.*
- *$f$ is $G$-Lipschitz continuous if $\forall x \in \mathbb{R}^d$, $\|\nabla f(x)\| \leq G$.*

### 2.1 Byzantine Convex Stochastic Optimization

We let $m$ be number of worker machines and assume at most an $\alpha$ fraction of them are Byzantine for $\alpha \in \left[0, \frac{1}{2}\right)$. We denote by $\mathsf{good} \subseteq [m]$ the set of good (i.e. non-Byzantine) machines, and the algorithm does not know $\mathsf{good}$.

We let $\mathcal{D}$ be a distribution over (not necessarily convex) functions $f_s \colon \mathbb{R}^d \to \mathbb{R}$. Our goal is to approximately minimize the following objective:

$$\min_{x \in \mathbb{R}^d} \left\{ f(x) \stackrel{\text{def}}{=} \mathbb{E}_{s \sim \mathcal{D}}[f_s(x)] \right\}, \tag{2.1}$$

where we assume $f$ is convex. In each iteration $k = 1, 2, \ldots, T$, the algorithm is allowed to specify a point $x_k$ and query $m$ machines. Each machine $i \in [m]$ gives back a vector $\nabla_{k,i} \in \mathbb{R}^d$ satisfying

**Assumption 2.2.** *For each iteration $k \in [T]$ and for every $i \in \mathsf{good}$, we have*
- *$\nabla_{k,i} = \nabla f_s(x_k)$ for a random sample $s \sim \mathcal{D}$, and*
- *$\|\nabla_{k,i} - \nabla f(x_k)\| \leq \mathcal{V}$.*

*Remark* 2.3. For each $k \in [T]$ and $i \notin \mathsf{good}$, the vector $\nabla_{k,i}$ can be adversarially chosen and may depend on $\{\nabla_{k',i}\}_{k' \leq k, i \in [m]}$. In particular, the Byzantine machines can even collude in an iteration.

## 2.2 Useful inequalities

The following classical concentration result will be useful for us:

**Lemma 2.4** (Pinelis' 1994 inequality [28])**.** *Let $X_1, \ldots, X_T \in \mathbb{R}^d$ be a random process satisfying $\mathbb{E}[X_t | X_1, \ldots, X_{t-1}] = 0$ and $\|X_t\| \leq M$. Then, $\Pr\left[\|X_1 + \cdots + X_T\|^2 > 2\log(2/\delta) M^2 T\right] \leq \delta$.*

The next fact is completely classical (for projected mirror descent).

**Fact 2.5.** *if $x_{k+1} = \arg\min_{y:\, \|y-x_1\| \leq D}\{\frac{1}{2}\|y - x_k\|^2 + \eta \langle \xi, y - x_k \rangle\}$, then $\forall u\colon \|u - x_1\| \leq D$, we have*

$$\langle \xi, x_k - u \rangle \leq \langle \xi, x_k - x_{k+1} \rangle - \frac{\|x_k - x_{k+1}\|^2}{2\eta} + \frac{\|x_k - u\|^2}{2\eta} - \frac{\|x_{k+1} - u\|^2}{2\eta} \ .$$

*Proof.* We have

$$\langle \xi, x_k - u \rangle = \langle \xi, x_k - x_{k+1} \rangle + \langle \xi, x_{k+1} - u \rangle \overset{\text{\textcircled{\scriptsize 1}}}{\leq} \langle \xi, x_k - x_{k+1} \rangle + \frac{1}{\eta}\langle x_k - x_{k+1}, x_{k+1} - u \rangle$$

$$= \langle \xi, x_k - x_{k+1} \rangle + \frac{\|x_k - u\|^2}{2\eta} - \frac{\|x_{k+1} - u\|^2}{2\eta} - \frac{\|x_k - x_{k+1}\|^2}{2\eta}$$

where inequality ① is by the constrained minimality of $x_{k+1}$ which implies for every $\|u - x_1\| \leq D$,

$$\langle x_{k+1} - x_k + \eta \xi, x_{k+1} - u \rangle \leq 0 \ . \qquad \square$$

## 3 ByzantineSGD for Non-Strongly Convex Objectives

Without loss of generality, in this section we assume that we are given a starting point $x_1 \in \mathbb{R}^d$ and want to solve the following more general problem:[4]

$$\min_{\|x-x_1\| \leq D} \left\{ f(x) \overset{\text{def}}{=} \mathbb{E}_{s \sim \mathcal{D}}[f_s(x)] \right\} \ . \tag{3.1}$$

We denote by $x^*$ an arbitrary minimizer to Problem (3.1).

Our algorithm ByzantineSGD is formally stated in Algorithm 1. In each iteration at point $x_k$, ByzantineSGD tries to identify a set $\mathsf{good}_k$ of "candidate good" machines, and then perform stochastic gradient update only with respect to $\mathsf{good}_k \subseteq [m]$, by using direction $\xi_k \overset{\text{def}}{=} \frac{1}{m}\sum_{i \in \mathsf{good}_k} \nabla_{k,i}$.

The way $\mathsf{good}_k$ is maintained is by constructing two "estimation sequences". Namely, for each machine $i \in [m]$, we maintain a real value $A_i = \sum_{t=1}^{k} \langle \nabla_{t,i}, x_t - x_1 \rangle$ and a vector $B_i = \sum_{t=1}^{k} \nabla_{t,i}$. Then, we denote by $A_{\mathsf{med}}$ the median of $\{A_1, \ldots, A_m\}$ and $B_{\mathsf{med}}$ some "vector median" of $\{B_1, \ldots, B_m\}$. We also define $\nabla_{\mathsf{med}}$ to be some "vector median" of $\{\nabla_{k,1}, \ldots, \nabla_{k,m}\}$.[5]

Starting from $\mathsf{good}_0 = [m]$, we define $\mathsf{good}_k$ to be all the machines $i$ from $\mathsf{good}_{k-1}$ whose $A_i$ is $\mathfrak{T}_A$-close to $A_{\mathsf{med}}$, $B_i$ is $\mathfrak{T}_B$-close to $B_{\mathsf{med}}$, and $\Delta_i$ is $4\mathcal{V}$-close to $\Delta_{\mathsf{med}}$. One can prove later that if the thresholds $\mathfrak{T}_A$ and $\mathfrak{T}_B$ are chosen appropriately, then $\mathsf{good}_k$ always contains all machines in $\mathsf{good}$.

---

[4]This is so because even in unconstrained setting, classical SGD requires knowing an upper bound $D$ to $\|x_1 - x^*\|$ in order to choose the learning rate. We can thus add the constraint to the objective.

[5]For instance for $\{\nabla_{k,1}, \ldots, \nabla_{k,m}\}$, our vector median is defined as follows. We select $\nabla_{\mathsf{med}}$ to be any $\nabla_{k,i}$ as long as $|\{j \in [m]\colon \|\nabla_j - \nabla_i\| \leq 2\mathcal{V}\}| > m/2$. Such an index $i \in [m]$ can be efficiently computed because our later lemmas shall ensure that at least $(1-\alpha)m$ indices in $[m]$ are valid choices for $i$. Therefore, one can for instance guess a random index $i$ and verify whether it is valid. In expectation at most 2 guesses are needed.

**Algorithm 1** ByzantineSGD($\eta, x_1, D, T, \mathfrak{T}_A, \mathfrak{T}_B$)
───────────────────────────────────────────────────────────────
**Input:** learning rate $\eta > 0$, starting point $x_1 \in \mathbb{R}^d$, diameter $D > 0$, number of iterations $T$, thresholds $\mathfrak{T}_A, \mathfrak{T}_B > 0$;
$\diamond$ *theory suggests* $\mathfrak{T}_A = 4D\mathcal{V}\sqrt{T\log(16mT/\delta)}$ *and* $\mathfrak{T}_B = 4\mathcal{V}\sqrt{T\log(16mT/\delta)}$

1: $\mathsf{good}_1 \leftarrow [m]$;
2: **for** $k \leftarrow 1$ **to** $T$ **do**
3:     **for** $i \leftarrow 1$ **to** $m$ **do**
4:         receive $\nabla_{k,i} \in \mathbb{R}^d$ from machine $i \in [m]$;     $\diamond$ *we have* $\mathbb{E}[\nabla_{k,i}] = \nabla f(x_k)$ *if* $i \in \mathsf{good}$
5:         $A_i \leftarrow \sum_{t=1}^{k}\langle \nabla_{t,i}, x_t - x_1\rangle$ and $B_i \leftarrow \sum_{t=1}^{k}\nabla_{t,i}$;
6:     **end for**
7:     $A_{\mathsf{med}} \stackrel{\mathrm{def}}{=} \mathsf{median}\{A_1, \ldots, A_m\}$
8:     $B_{\mathsf{med}} \leftarrow B_i$ where $i \in [m]$ is any machine s.t. $\left|\{j \in [m] \colon \|B_j - B_i\| \leq \mathfrak{T}_B\}\right| > m/2$.
$\diamond$ *all machines* $i \in \mathsf{good}$ *will be valid choice, see Claim 3.2b*
9:     $\nabla_{\mathsf{med}} \leftarrow \nabla_{k,i}$ where $i \in [m]$ is any machine s.t. $\left|\{j \in [m] \colon \|\nabla_{k,j} - \nabla_{k,i}\| \leq 2\mathcal{V}\}\right| > m/2$
$\diamond$ *all machines* $i \in \mathsf{good}$ *will be valid choice due to Assumption 2.2*
10:     $\mathsf{good}_k \leftarrow \{i \in \mathsf{good}_{k-1} \colon |A_i - A_{\mathsf{med}}| \leq \mathfrak{T}_A \wedge \|B_i - B_{\mathsf{med}}\| \leq \mathfrak{T}_B \wedge \|\nabla_{k,i} - \nabla_{\mathsf{med}}\| \leq 4\mathcal{V}\}$;
$\diamond$ *with high probability* $\mathsf{good}_k \supseteq \mathsf{good}$
11:     $x_{k+1} = \arg\min_{y \colon \|y - x_1\| \leq D}\left\{\frac{1}{2}\|y - x_k\|^2 + \eta\langle \frac{1}{m}\sum_{i \in \mathsf{good}_k}\nabla_{k,i}, y - x_k\rangle\right\}$;
12: **end for**
───────────────────────────────────────────────────────────────

**High-Level Idea.** As we shall see, the "error" incurred by ByzantineSGD contains two parts:

$$\mathsf{Error}_1 \stackrel{\mathrm{def}}{=} \sum_{k \in [T]}\sum_{i \in \mathsf{good}_k}\langle \nabla_{k,i} - \nabla f(x_k), x_k - x^*\rangle$$

$$\mathsf{Error}_2 \stackrel{\mathrm{def}}{=} \frac{1}{T}\sum_{k \in [T]}\left\|\frac{1}{m}\sum_{i \in \mathsf{good}_k}(\nabla_{k,i} - \nabla f(x_k))\right\|^2$$

Intuitively, $\mathsf{Error}_1$ is due to the bias created by the stochastic gradient (of good machines) and the adversarial noise (of Byzantine machines); while $\mathsf{Error}_2$ is the variance of using $\xi_k$ to approximate $\nabla f(x_k)$.

As we shall see, $\mathsf{Error}_2$ is almost always "well bounded." However, the adversarial noise incurred in $\mathsf{Error}_1$ can sometimes destroy the convergence of SGD. We therefore use $\{A_i\}_i$ and $\{B_i\}_i$ to perform a reasonable estimation of $\mathsf{Error}_1$, and remove the bad machines if they misbehave. Note that even at the end of the algorithm, $\mathsf{good}_T$ may still contain some Byzantine machines; however, their adversarial noise must be negligible and shall not impact the performance of the algorithm.

### 3.1 Concentration Lemmas

We first prove some concentration inequalities. Let us denote by $\nabla_k \stackrel{\mathrm{def}}{=} \nabla f(x_k)$ and $C \stackrel{\mathrm{def}}{=} \log(16mT/\delta)$. We consider three events, all of them happen with high probability.

**Event A.** Denote by

$$A_i^{(t)} \stackrel{\mathrm{def}}{=} \sum_{k=1}^{t}\langle \nabla_{k,i}, x_k - x_1\rangle \;, \quad A_\star^{(t)} \stackrel{\mathrm{def}}{=} \sum_{k=1}^{t}\langle \nabla_k, x_k - x_1\rangle \quad \text{and} \quad A_{\mathsf{med}}^{(t)} \stackrel{\mathrm{def}}{=} \mathsf{median}\{A_1^{(t)}, \ldots, A_m^{(t)}\}$$

Recall that the algorithm computes $A_i^{(t)}$ and $A_{\mathsf{med}}^{(t)}$ for all $i \in [m]$ and $t \in [T]$. In contrast, the value $A_\star^{(t)}$ is unknown.

**Claim 3.1.** *With probability at least $1 - \delta/4$, we have*

(a) *for all $i \in$ good and $t \in [T]$, $|A_i^{(t)} - A_\star^{(t)}| \leq 2D\mathcal{V}\sqrt{tC}$.*

(b) *for all $i \in$ good and $t \in [T]$, $|A_i^{(t)} - A_{\text{med}}^{(t)}| \leq 4D\mathcal{V}\sqrt{tC}$ and $|A_\star^{(t)} - A_{\text{med}}^{(t)}| \leq 2D\mathcal{V}\sqrt{tC}$.*

(c) *$|\sum_{i \in \text{good}}(A_i^{(T)} - A_\star^{(T)})| \leq 2D\mathcal{V}\sqrt{TmC}$.*

*We denote this event by $\mathsf{Event}_A$.*

*Proof.*

(a) Note that $\mathbb{E}[\langle \nabla_{k,i}, x_k - x_1 \rangle] = \langle \nabla_k, x_k - x_1 \rangle$ and $|\langle \nabla_{k,i} - \nabla_k, x_k - x_1 \rangle| \leq \|\nabla_{k,i} - \nabla_k\| \|x_k - x_1\| \leq \mathcal{V}D$. We can thus apply Lemma 2.4 with $X_k = \langle \nabla_{k,i} - \nabla_k, x_k - x_1 \rangle$ to derive that with probability at least $1 - \frac{1}{8mT}$ we have $|A_i^{(t)} - A_\star^{(t)}| \leq 2D\mathcal{V}\sqrt{tC}$. Claim 3.1a then follows by taking a union bound over $i \in$ good and $t \in [T]$.

(b) $|A_\star^{(t)} - A_{\text{med}}^{(t)}| \leq 2D\mathcal{V}\sqrt{tC}$ follows from Claim 3.1a as the fact that $|\text{good}| > m/2$. $|A_i^{(t)} - A_{\text{med}}^{(t)}| \leq 4D\mathcal{V}\sqrt{tC}$ then follows by triangle inequality.

(c) We can apply Lemma 2.4 with $\{X_1, X_2, \ldots, X_{T|\text{good}|}\} = \{\langle \nabla_{k,i} - \nabla_k, x_k - x_1 \rangle\}_{k \in [T], i \in \text{good}}$. □

**Event B.** Denote by

$$B_i^{(t)} \stackrel{\text{def}}{=} \sum_{k=1}^{t} \nabla_{k,i} \ , \quad B_\star^{(t)} \stackrel{\text{def}}{=} \sum_{k=1}^{t} \nabla_k$$

Recall for each $t \in [T]$, the algorithm computes $\{B_1^{(t)}, \ldots, B_m^{(t)}\}$ as well as some $B_{\text{med}}^{(t)} = B_i^{(t)}$ where $i$ is any machine in $[m]$ such that at least half of $j \in [m]$ satisfies $\|B_j^{(t)} - B_i^{(t)}\| \leq 4\mathcal{V}\sqrt{tC}$.

**Claim 3.2.** *With probability at least $1 - \delta/4$, we have*

(a) *for all $i \in$ good and $t \in [T]$, $\|B_i^{(t)} - B_\star^{(t)}\| \leq 2\mathcal{V}\sqrt{tC}$.*

(b) *for all $t \in [T]$, each $i \in$ good is a valid choice for $B_{\text{med}}^{(t)} = B_i^{(t)}$.*

(c) *for all $i \in$ good and $t \in [T]$, $\|B_i^{(t)} - B_{\text{med}}^{(t)}\| \leq 4\mathcal{V}\sqrt{tC}$ and $\|B_\star^{(t)} - B_{\text{med}}^{(t)}\| \leq 6\mathcal{V}\sqrt{tC}$*

(d) *$\|\sum_{i \in \text{good}}(B_i^{(T)} - B_\star^{(T)})\| \leq 2\mathcal{V}\sqrt{TmC}$.*

*We denote this event by $\mathsf{Event}_B$.*

*Proof.*

(a) Note that $\mathbb{E}[\nabla_{k,i}] = \nabla_k$ and $\|\nabla_{k,i} - \nabla_k\| \leq \mathcal{V}$. We can thus apply Lemma 2.4 with $X_k = \nabla_{k,i} - \nabla_k$ and then take a union bound over all $i \in$ good and $t \in [T]$.

(b) Claim 3.2a implies for every $i, j \in$ good we have $\|B_i^{(t)} - B_j^{(t)}\| \leq 4\mathcal{V}\sqrt{tC}$. Therefore each $i \in$ good is a valid choice for setting $B_{\text{med}}^{(t)} = B_i^{(t)}$.

(c) The is a consequence of Claim 3.2a and the definition of $B_{\text{med}}^{(t)}$.

(d) We can apply Lemma 2.4 with $\{X_1, X_2, \ldots, X_{T|\text{good}|}\} = \{\nabla_{k,i} - \nabla_k\}_{k \in [T], i \in \text{good}}$. □

**Event C.**

**Claim 3.3.** *With probability at least $1 - \delta/4$, we have $\left\|\frac{1}{m}\sum_{i \in \text{good}}(\nabla_{k,i} - \nabla_k)\right\|^2 \leq 2\frac{\mathcal{V}^2}{m}C$ for all $k \in [T]$. We denote this event by $\mathsf{Event}_C$.*

*Proof.* For each $k \in [T]$, one can define $X_j = \nabla_{k,j} - \nabla_k$ for $j \in \mathsf{good}$ and apply Lemma 2.4. The claim then follows after taking union bound over $k \in [T]$. □

## 3.2 Bounding Error Terms

In this subsection we bound the two error terms $\mathsf{Error}_1$ and $\mathsf{Error}_2$. Recall that the thresholding parameters $\mathfrak{T}_A = 4D\mathcal{V}\sqrt{TC}$ and $\mathfrak{T}_B = 4\mathcal{V}\sqrt{TC}$, and for all $k \in [T]$, let $\nabla_{\mathsf{med}}^{(k)}$ be the $\nabla_{\mathsf{med}}$ computed in iteration $k$. We make two simple observations:

**Claim 3.4.** *For all $k \in [T]$, each $i \in \mathsf{good}$ is a valid choice for $\nabla_{\mathsf{med}}^{(k)} = \nabla_{k,i}$ and $\|\nabla_{\mathsf{med}}^{(k)} - \nabla_k\| \leq 3\mathcal{V}$.*

*Proof.* Observe that for any $i, j \in \mathsf{good}$ we have $\|\nabla_{k,i} - \nabla_{k,j}\|_2 \leq 2\mathcal{V}$. Therefore, given $\alpha < 1/2$, every $i \in \mathsf{good}$ is a valid choice for selecting $\nabla_{\mathsf{med}}^{(k)} = \nabla_{k,i}$. As for the second part, if $\|\nabla_{\mathsf{med}}^{(k)} - \nabla_k\| > 3\mathcal{V}$, by reverse triangle inequality this implies $\|\nabla_{\mathsf{med}}^{(k)} - \nabla_{k,i}\| > 2\mathcal{V}$ for all $i \in \mathsf{good}$, which contradicts the definition of $\nabla_{\mathsf{med}}^{(k)}$ because $\alpha < 1/2$. □

**Claim 3.5.** *Under $\mathsf{Event}_A$ and $\mathsf{Event}_B$, we have $\mathsf{good}_k \supseteq \mathsf{good}$ for all $k \in [T]$.*

*Proof.* Observe that Claim 3.1b, Claim 3.2c and Claim 3.4 together imply that good machines $i \in \mathsf{good}$ always satisfy $|A_i^{(k)} - A_{\mathsf{med}}^{(k)}| \leq \mathfrak{T}_A$, $\|B_i^{(k)} - B_{\mathsf{med}}^{(k)}\| \leq \mathfrak{T}_B$ and $\|\nabla_{\mathsf{med}}^{(k)} - \nabla_{k,i}\| \leq 4\mathcal{V}$. Thus no elements from $\mathsf{good}$ will ever be removed from $\mathsf{good}_k$ in all iterations $k = 1, 2, \ldots, T$. □

**Lemma 3.6.** *If $\mathsf{Event}_A$ and $\mathsf{Event}_B$ hold, we have*
$$|\mathsf{Error}_1| \leq 4D\mathcal{V}\sqrt{TmC} + 16\alpha m D\mathcal{V}\sqrt{TC}$$

*Proof.* For each machine $i \in [m]$, denote by $T_i \in \{0, 1, \ldots, T\}$ the maximum iteration index so that $i \in \mathsf{good}_{T_i}$. We calculate that
$$\sum_{k \in [T]} \sum_{i \in \mathsf{good}_k} \langle \nabla_{k,i} - \nabla_k, x_k - x^* \rangle = \sum_{i \in \mathsf{good}} \left(A_i^{(T)} - A_\star^{(T)} + \langle B_i^{(T)} - B_\star^{(T)}, x_0 - x^* \rangle\right)$$
$$+ \sum_{i \notin \mathsf{good}} \left(A_i^{(T_i)} - A_\star^{(T_i)} + \langle B_i^{(T_i)} - B_\star^{(T_i)}, x_0 - x^* \rangle\right) \quad (3.2)$$

As for the first summation on the right hand side of (3.2), according to Claim 3.1c and Claim 3.2d respectively, we have $|\sum_{i \in \mathsf{good}} (A_i^{(T)} - A_\star^{(T)})| \leq 2D\mathcal{V}\sqrt{TmC}$ and $\|\sum_{i \in \mathsf{good}} \langle B_i^{(T)} - B_\star^{(T)}, x_0 - x^* \rangle\| \leq 2D\mathcal{V}\sqrt{TmC}$.

As for the second summation on the right hand side of (3.2), we focus on each $i \notin \mathsf{good}$. By the definition of $\mathfrak{T}_A$ and $T_i$, we have $|A_i^{(T_i)} - A_{\mathsf{med}}^{(T_i)}| \leq \mathfrak{T}_A = 4D\mathcal{V}\sqrt{T_iC}$ and therefore (using Claim 3.1b)
$$|A_i^{(T_i)} - A_\star^{(T_i)}| \leq |A_i^{(T_i)} - A_{\mathsf{med}}^{(T_i)}| + |A_{\mathsf{med}}^{(T_i)} - A_\star^{(T_i)}| \leq 6D\mathcal{V}\sqrt{TC} \ .$$

Similarly, for each $i \notin \mathsf{good}$, we have by the definition of $\mathfrak{T}_B$ and $T_i$, we have $\|B_i^{(T_i)} - B_{\mathsf{med}}^{(T_i)}\| \leq \mathfrak{T}_B = 4\mathcal{V}\sqrt{T_iC}$ and therefore (using Claim 3.2c)
$$\|B_i^{(T_i)} - B_\star^{(T_i)}\| \leq \|B_i^{(T_i)} - B_{\mathsf{med}}^{(T_i)}\| + \|B_{\mathsf{med}}^{(T_i)} - B_\star^{(T_i)}\| \leq 10\mathcal{V}\sqrt{TC} \ .$$

Finally, putting everything back to (3.2), and using the fact that $|[m] \setminus \mathsf{good}| = \alpha m$, we finish the proof of Lemma 3.6. □

**Lemma 3.7.** *If $\mathsf{Event}_A$, $\mathsf{Event}_B$ and $\mathsf{Event}_C$ hold, we have $\mathsf{Error}_2 \leq 32\alpha^2\mathcal{V}^2 + \frac{4\mathcal{V}^2 C}{m}$ .*

*Proof.* We have for each $k \in [T]$,

$$\left\|\frac{1}{m}\sum_{i\in \mathsf{good}_k}(\nabla_{k,i} - \nabla_k)\right\|^2 \leq 2\left\|\frac{1}{m}\sum_{i\in \mathsf{good}}(\nabla_{k,i} - \nabla_k)\right\|^2 + 2\left\|\frac{1}{m}\sum_{i\in \mathsf{good}_k\setminus\mathsf{good}}(\nabla_{k,i} - \nabla_k)\right\|^2$$

$$\leq \frac{4\mathcal{V}^2 C}{m} + 32\alpha^2 \mathcal{V}^2 \ .$$

Above, the first inequality is because $\|a+b\|^2 \leq 2\|a\|^2 + 2\|b\|^2$, and the second inequality uses Claim 3.3, the fact that $|\mathsf{good}_k \setminus \mathsf{good}| \leq m - |\mathsf{good}| = \alpha m$, and Claim 3.4. $\square$

## 3.3 Smooth Case

**Theorem 3.8.** *Suppose in Problem (3.1) our $f(x)$ is L-smooth and Assumption 2.2 holds. Suppose $\eta \leq \frac{1}{2L}$ and $\mathfrak{T}_A = 4D\mathcal{V}\sqrt{TC}$ and $\mathfrak{T}_B = 4\mathcal{V}\sqrt{TC}$. Then, with probability at least $1-\delta$, letting $C \stackrel{\text{def}}{=} \log(16mT/\delta)$ and $\overline{x} \stackrel{\text{def}}{=} \frac{x_2+\cdots+x_{T+1}}{T}$, we have*

$$f(\overline{x}) - f(x^*) \leq \frac{D^2}{\eta T} + \frac{8D\mathcal{V}\sqrt{TmC} + 32\alpha m D\mathcal{V}\sqrt{TC}}{Tm} + \eta \cdot \left(\frac{8\mathcal{V}^2 C}{m} + 32\alpha^2 \mathcal{V}^2\right) \ .$$

*If $\eta$ is chosen optimally, then*

$$f(\overline{x}) - f(x^*) \leq O\left(\frac{LD^2}{T} + \frac{D\mathcal{V}\sqrt{C}}{\sqrt{Tm}} + \frac{\alpha D\mathcal{V}\sqrt{C}}{\sqrt{T}}\right) \ .$$

We remark that

- The first term $O\left(\frac{LD^2}{T}\right)$ is the classical error rate for gradient descent on smooth objectives [26] and should exist even if $\mathcal{V} = 0$ (so every $\nabla_{k,i}$ exactly equals $\nabla f(x_k)$) and $\alpha = 0$.
- The first two terms $\widetilde{O}\left(\frac{LD^2}{T} + \frac{D\mathcal{V}}{\sqrt{Tm}}\right)$ together match the classical mini-batch error rate for SGD on smooth objectives, and should exist even if $\alpha = 0$ (so we have no Byzantine machines).
- The third term $\widetilde{O}\left(\frac{\alpha D\mathcal{V}}{\sqrt{T}}\right)$ is optimal in our Byzantine setting due to Theorem 5.4.

*Proof of Theorem 3.8.* Applying Fact 2.5 for $k = 1, 2, \ldots, T$ with $u = x^*$, we have

$$\frac{1}{T}\sum_{k\in[T]}\langle \xi_k, x_k - x^*\rangle \leq \frac{D^2}{2\eta T} + \frac{1}{T}\sum_{k\in[T]}\left(\langle \xi_k, x_k - x_{k+1}\rangle - \frac{1}{2\eta}\|x_k - x_{k+1}\|^2\right)$$

$$= \frac{D^2}{2\eta T} + \frac{1}{T}\sum_{k\in[T]}\left(\left\langle\frac{1}{m}\sum_{i\in \mathsf{good}_k}\nabla_{k,i}, x_k - x_{k+1}\right\rangle - \frac{1}{2\eta}\|x_k - x_{k+1}\|^2\right) \quad (3.3)$$

We notice that the left hand side of (3.3)

$$\sum_{k\in[T]}\langle \xi_k, x_k - x^*\rangle = \frac{1}{m}\sum_{k\in[T]}\sum_{i\in\mathsf{good}_k}\langle\nabla_k, x_k - x^*\rangle + \frac{1}{m}\sum_{k\in[T]}\sum_{i\in\mathsf{good}_k}\langle\nabla_{k,i} - \nabla_k, x_k - x^*\rangle$$

$$\overset{\text{①}}{\geq} \frac{1}{m}\sum_{k\in[T]}\sum_{i\in\mathsf{good}_k}\left(f(x_k) - f(x^*)\right) + \frac{\mathsf{Error}_1}{m}$$

$$\overset{\text{②}}{\geq} \frac{1}{m}\sum_{k\in[T]}\sum_{i\in\mathsf{good}_k}\left(f(x_{k+1}) - f(x^*) - \langle\nabla_k, x_{k+1} - x_k\rangle - \frac{L}{2}\|x_k - x_{k+1}\|^2\right) + \frac{\mathsf{Error}_1}{m}$$

(3.4)

Above, inequality ① uses the convexity of $f(\cdot)$ and the definition of $\mathsf{Error}_1$, and inequality ② uses the smoothness of $f(\cdot)$ which implies $f(x_{k+1}) \leq f(x_k) + \langle \nabla f(x_k), x_{k+1} - x_k \rangle + \frac{L}{2}\|x_k - x_{k+1}\|^2$.

Putting (3.4) back to (3.3), we have

$$\frac{1}{Tm}\sum_{k\in[T]}\sum_{i\in\mathsf{good}_k}\big(f(x_{k+1}) - f(x^*)\big)$$

$$\leq \frac{D^2}{2\eta T} - \frac{\mathsf{Error}_1}{Tm} + \frac{1}{T}\sum_{k\in[T]}\left(\Big\langle \frac{1}{m}\sum_{i\in\mathsf{good}_k}(\nabla_{k,i} - \nabla_k), x_k - x_{k+1}\Big\rangle - \Big(\frac{1}{2\eta} - \frac{L}{2}\Big)\|x_k - x_{k+1}\|^2\right)$$

$$\overset{①}{\leq} \frac{D^2}{2\eta T} - \frac{\mathsf{Error}_1}{Tm} + \frac{\eta}{T}\sum_{k\in[T]}\Big\|\frac{1}{m}\sum_{i\in\mathsf{good}_k}(\nabla_{k,i} - \nabla_k)\Big\|^2 = \frac{D^2}{2\eta T} - \frac{\mathsf{Error}_1}{Tm} + \eta\mathsf{Error}_2 \ . \qquad (3.5)$$

Above, inequality ① uses the fact that $\frac{1}{2\eta} - \frac{L}{2} \geq \frac{1}{4\eta}$, and Young's inequality which says $\langle a, b\rangle - \frac{1}{2}\|b\|^2 \leq \frac{1}{2}\|a\|^2$.

Finally, we conclude the proof by plugging Lemma 3.6, Lemma 3.7 and the following convexity inequality into (3.5):

$$\frac{1}{Tm}\sum_{k\in[T]}\sum_{i\in\mathsf{good}_k}\big(f(x_{k+1}) - f(x^*)\big) = \frac{1}{T}\sum_{k\in[T]}\frac{|\mathsf{good}_k|}{m}\big(f(x_k) - f(x^*)\big)$$

$$\geq \frac{1}{T}\sum_{k\in[T]}\frac{1}{2}\big(f(x_k) - f(x^*)\big) \geq \frac{1}{2}\big(f(\bar{x}) - f(x^*)\big) \ . \qquad \square$$

## 3.4 Nonsmooth Case

**Theorem 3.9.** *Suppose in Problem (3.1) our $f(x)$ is differentiable, $G$-Lipschitz continuous and Assumption 2.2 holds. Suppose $\eta > 0$ and $\mathfrak{T}_A = 4D\mathcal{V}\sqrt{TC}$ and $\mathfrak{T}_B = 4\mathcal{V}\sqrt{TC}$. Then, with probability at least $1 - \delta$, letting $C \overset{\text{def}}{=} \log(16mT/\delta)$ and $\bar{x} \overset{\text{def}}{=} \frac{x_1 + \cdots + x_T}{T}$, we have*

$$f(\bar{x}) - f(x^*) \leq \frac{D^2}{\eta T} + \frac{2\eta G^2}{T} + \frac{8D\mathcal{V}\sqrt{TmC} + 32\alpha m D\mathcal{V}\sqrt{TC}}{Tm} + \eta \cdot \Big(\frac{8\mathcal{V}^2 C}{m} + 32\alpha^2\mathcal{V}^2\Big) \ .$$

*If $\eta$ is chosen optimally, then*

$$f(\bar{x}) - f(x^*) \leq O\Big(\frac{GD}{\sqrt{T}} + \frac{D\mathcal{V}\sqrt{C}}{\sqrt{Tm}} + \frac{\alpha D\mathcal{V}\sqrt{C}}{\sqrt{T}}\Big) \ .$$

We remark that

- The first term $O\big(\frac{GD}{\sqrt{T}}\big)$ is the classical error rate for gradient descent on non-smooth objectives [5], and should exist even if $\mathcal{V} = 0$ (so every $\nabla_{k,i}$ exactly equals $\nabla f(x_k)$) and $\alpha = 0$.

- The first two terms $\widetilde{O}\big(\frac{GD}{\sqrt{T}} + \frac{D\mathcal{V}}{\sqrt{Tm}}\big)$ together match the classical mini-batch error rate for SGD on non-smooth objectives, and should exist even if $\alpha = 0$ (so we have no Byzantine machines).

- The third term $\widetilde{O}\big(\frac{\alpha D\mathcal{V}}{\sqrt{T}}\big)$ is optimal in our Byzantine setting due to Theorem 5.4.

*Proof of Theorem 3.9.* Applying Fact 2.5 for $k = 1, 2, \ldots, T$ with $u = x^*$, we have

$$\frac{1}{T} \sum_{k \in [T]} \langle \xi_k, x_k - x^* \rangle \leq \frac{D^2}{2\eta T} + \frac{1}{T} \sum_{k \in [T]} \left( \langle \xi_k, x_k - x_{k+1} \rangle - \frac{1}{2\eta} \|x_k - x_{k+1}\|^2 \right)$$

$$\stackrel{①}{\leq} \frac{D^2}{2\eta T} + \frac{\eta}{2T} \sum_{k \in [T]} \|\xi_k\|^2 = \frac{D^2}{2\eta T} + \frac{\eta}{2T} \sum_{k \in [T]} \left\| \frac{1}{m} \sum_{i \in \mathsf{good}_k} \nabla_{k,i} \right\|^2$$

$$\stackrel{②}{\leq} \frac{D^2}{2\eta T} + \frac{\eta}{2T} \sum_{k \in [T]} \left( 2\|\nabla_k\|^2 + 2 \left\| \frac{1}{m} \sum_{i \in \mathsf{good}_k} (\nabla_{k,i} - \nabla_k) \right\|^2 \right)$$

$$\stackrel{③}{\leq} \frac{D^2}{2\eta T} + \frac{\eta G^2}{T} + \eta \mathsf{Error}_2 \ . \tag{3.6}$$

Above, inequality ① uses Young's inequality $\langle a, b \rangle - \frac{1}{2}\|b\|^2 \leq \frac{1}{2}\|a\|^2$; inequality ② uses $\|a+b\|^2 \leq 2\|a\|^2 + 2\|b\|^2$; and inequality ③ uses the $G$-Lipschitz continuity of $f(\cdot)$ and the definition of $\mathsf{Error}_2$.

We notice that the left hand side of (3.6)

$$\sum_{k \in [T]} \langle \xi_k, x_k - x^* \rangle = \frac{1}{m} \sum_{k \in [T]} \sum_{i \in \mathsf{good}_k} \langle \nabla_k, x_k - x^* \rangle + \frac{1}{m} \sum_{k \in [T]} \sum_{i \in \mathsf{good}_k} \langle \nabla_{k,i} - \nabla_k, x_k - x^* \rangle$$

$$\stackrel{①}{\geq} \frac{1}{m} \sum_{k \in [T]} \sum_{i \in \mathsf{good}_k} \left( f(x_k) - f(x^*) \right) + \frac{\mathsf{Error}_1}{m}$$

$$= \sum_{k \in [T]} \frac{|\mathsf{good}_k|}{m} \left( f(x_k) - f(x^*) \right) + \frac{\mathsf{Error}_1}{m}$$

$$\stackrel{②}{\geq} \sum_{k \in [T]} \frac{1}{2} \left( f(x_k) - f(x^*) \right) + \frac{\mathsf{Error}_1}{m} \stackrel{③}{\geq} \frac{T}{2} \left( f(\overline{x}) - f(x^*) \right) + \frac{\mathsf{Error}_1}{m} \tag{3.7}$$

Above, inequality ① uses the convexity of $f(\cdot)$ and the definition of $\mathsf{Error}_1$, inequality ② uses Claim 3.3 which implies $|\mathsf{good}_k| = (1-\alpha)m \geq \frac{m}{2}$, and inequality ③ uses the convexity of $f(\cdot)$ again.

Putting (3.7) back to (3.6), and plugging in Lemma 3.6 and Lemma 3.7 we finish the proof. □

## 4 ByzantineSGD in Strongly Convex Objectives

In this section we solve[6]

$$\min_{x \in \mathbb{R}^d} \left\{ f(x) \stackrel{\text{def}}{=} \mathbb{E}_{s \sim \mathcal{D}}[f_s(x)] \right\} \quad \text{where } f(x) \text{ is } \sigma\text{-strongly convex.} \tag{4.1}$$

We denote by $x^*$ the minimizer to Problem (4.1). We propose an epoch-based method which repeatedly applies `ByzantineSGD` a logarithmic number of times in order to minimize Problem (4.1).

*Remark 4.1.* Traditionally, to achieve a high-confidence result of SGD for the strongly convex objectives, one either applies a reduction to non-strongly convex ones (c.f. Hazan and Kale [17]), or a sophisticated martingale analysis for a single-run algorithm (c.f. Rakhlin et al. [30]). We choose to use the reduction approach for presenting the simplest analysis.

Suppose we are given an initial vector $x_0$ with some upper bound $D > 0$ satisfying $\|x^{(0)} - x^*\| \leq D$. Define $D_0 = D$. Our algorithm will consists of $P$ epochs.

---

[6]To present the simplest result, we have assumed that Problem (4.1) is unconstrained. One can also impose an addition constraint $\|x - x_0\| \leq D$ but we refrain from doing so.

In each epoch $p = 1, 2, \ldots, P$, starting at $x^{(p-1)}$, we want to find a point $x$ satisfying $\|x^{(p)} - x^*\| \leq D_p \overset{\text{def}}{=} 2^{-p} D$. To achieve so, it suffices to find $x^{(p)}$ satisfying $f(x^{(p)}) - f(x^*) \leq \frac{\sigma D_p^2}{2}$, because we always have $\frac{\sigma}{2} \|x - x^*\|^2 \leq f(x) - f(x^*)$ as a consequence of strong convexity. We choose $P = \lceil \log_2 \sqrt{\frac{\sigma D^2}{2\varepsilon}} \rceil$ so that we have $f(x^{(P)}) - f(x^*) \leq \frac{\sigma D_P^2}{2} \leq \varepsilon$ after the last epoch is done.

This epoch-based algorithm gives rise to the following two theorems:

**Theorem 4.2.** *Suppose in Problem (4.1) our $f(x)$ is L-smooth and Assumption 2.2 holds. Given $x_0 \in \mathbb{R}^d$ with guarantee $\|x_0 - x^*\| \leq D$, one can repeatedly apply* `ByzantineSGD` *to find a point $x$ satisfying with probability at least $1 - \delta'$,*

$$f(x) - f(x^*) \leq \varepsilon \quad \text{and} \quad \|x - x^*\|^2 \leq 2\varepsilon/\sigma$$

*in*

$$T = \widetilde{O}\Big(\frac{L}{\sigma} + \frac{\mathcal{V}^2}{m\sigma\varepsilon} + \frac{\alpha^2 \mathcal{V}^2}{\sigma\varepsilon}\Big)$$

*iterations, where the $\widetilde{O}$ notation hides logarithmic factors in $D, m, L, \mathcal{V}, \sigma^{-1}, \varepsilon^{-1}, \delta^{-1}$.*

We remark that

- The first term $\widetilde{O}\big(\frac{L}{\sigma}\big)$ is the classical for gradient descent on smooth and strongly convex objectives [26], and should exist even if $\mathcal{V} = 0$ (so every $\nabla_{k,i}$ exactly equals $\nabla f(x_k)$) and $\alpha = 0$.
- The first two terms $\widetilde{O}\big(\frac{L}{\sigma} + \frac{\mathcal{V}^2}{m\sigma\varepsilon}\big)$ together match the classical mini-batch error rate for SGD on smooth and strongly convex objectives, and should exist even if $\alpha = 0$.
- The third term $\widetilde{O}\big(\frac{\alpha^2 \mathcal{V}^2}{\sigma\varepsilon}\big)$ is optimal in our Byzantine setting due to Theorem 5.4.

*Proof of Theorem 4.2.* Starting at $x^{(0)}$, in each epoch $p$, we start from $x^{(p-1)}$ with the guarantee that $\|x^{(p-1)} - x^*\| \leq D_{p-1}$, and wish to find a point $x^{(p)}$ satisfying $f(x^{(p)}) - f(x^*) \leq \frac{\sigma D_p^2}{2}$. We can apply Theorem 3.8 for this purpose, and let $T_p$ be the number of iterations needed. We have

$$T_p = O\Big(\frac{L}{\sigma} + \frac{\mathcal{V}^2 C}{m\sigma^2 D_p^2} + \frac{\alpha^2 \mathcal{V}^2 C}{\sigma^2 D_p^2}\Big) \ .$$

Finally, summing up $T = T_1 + \cdots + T_P$ and using $D_P^2 = \Theta(\varepsilon/\sigma)$ give the desired result. □

**Theorem 4.3.** *Suppose in Problem (4.1) our $f(x)$ is differentiable, G-Lipschitz continuous and Assumption 2.2 holds. Given $x_0 \in \mathbb{R}^d$ with guarantee $\|x_0 - x^*\| \leq D$, one can repeatedly apply* `ByzantineSGD` *to find a point $x$ satisfying with probability at least $1 - \delta'$,*

$$f(x) - f(x^*) \leq \varepsilon \quad \text{and} \quad \|x - x^*\|^2 \leq 2\varepsilon/\sigma$$

*in*

$$T = \widetilde{O}\Big(\frac{G^2}{\sigma\varepsilon} + \frac{\mathcal{V}^2}{m\sigma\varepsilon} + \frac{\alpha^2 \mathcal{V}^2}{\sigma\varepsilon} + 1\Big)$$

*iterations, where the $\widetilde{O}$ notation hides logarithmic factors in $D, m, L, \mathcal{V}, \sigma^{-1}, \varepsilon^{-1}, \delta^{-1}$.*

We remark that

- The first term $\widetilde{O}\big(\frac{G^2}{\sigma\varepsilon}\big)$ is the classical for gradient descent on non-smooth and strongly convex objectives [5], and should exist even if $\mathcal{V} = 0$ (so every $\nabla_{k,i}$ exactly equals $\nabla f(x_k)$) and $\alpha = 0$.

- The first two terms $\widetilde{O}\big(\frac{G^2}{\sigma\varepsilon} + \frac{\mathcal{V}^2}{m\sigma\varepsilon}\big)$ together match the classical mini-batch error rate for SGD on non-smooth and strongly convex objectives, and should exist even if $\alpha = 0$.
- The third term $\widetilde{O}\big(\frac{\alpha^2\mathcal{V}^2}{\sigma\varepsilon}\big)$ is optimal in our Byzantine setting due to Theorem 5.4.

*Proof of Theorem 4.3.* Starting at $x^{(0)}$, in each epoch $p$, we start from $x^{(p-1)}$ with the guarantee that $\|x^{(p-1)} - x^*\| \leq D_{p-1}$, and wish to find a point $x^{(p)}$ satisfying $f(x^{(p)}) - f(x^*) \leq \frac{\sigma D_p^2}{2}$. We can apply Theorem 3.9 for this purpose, and let $T_p$ be the number of iterations needed. We have

$$T_p = O\Big(\frac{G^2}{\sigma^2 D_p^2} + \frac{\mathcal{V}^2 C}{m\sigma^2 D_p^2} + \frac{\alpha^2 \mathcal{V}^2 C}{\sigma^2 D_p^2}\Big) \ .$$

Finally, summing up $T = T_1 + \cdots + T_P$ and using $D_P^2 = \Theta(\varepsilon/\sigma)$ give the desired result. $\square$

## 5 Lower Bounds in Byzantine Stochastic Optimization

In this section, we prove that the convergence rates we obtain in Sections 3 and 4 are optimal up to log factors, even in $d = 1$ dimension. Recall a random vector $X \in \mathbb{R}^d$ is subgaussian with variance proxy $\mathcal{V}^2$ if $u^T X$ is a univariate subgaussian random variable with variance proxy $\mathcal{V}^2$ for all unit vectors $u \in \mathbb{R}^d$. We require the following definition:

**Definition 5.1** (Stochastic estimator). *Given $\mathcal{X} \subseteq \mathbb{R}^d$ and $f \colon \mathcal{X} \to \mathbb{R}$, we say a random function $f_s$ (with $s$ drawn from some distribution $\mathcal{D}$) is a* stochastic estimator *for $f$ if $\mathbb{E}[f_s(x)] = f(x)$ for all $x \in \mathcal{X}$. Furthermore, we say $f_s$ is* subgaussian with variance proxy $\mathcal{V}^2$ *if $\nabla f_s(x) - \nabla f(x)$ is a subgaussian random variable with variance proxy $\mathcal{V}^2/d$ for all $x \in \mathcal{X}$.*

Note that the normalization factor of $1/d$ in this definition ensures that $\mathbb{E}\left[\|\nabla f_s(x) - \nabla f(x)\|^2\right] \leq O(\mathcal{V}^2)$, which matches the normalization used in this paper and throughout the literature. However, in our lower bound constructions it turns out that it suffices to take $d = 1$.

We prove our lower bounds only against subgaussian stochastic estimators. This is different from our Assumption 2.2 used in the upper-bound theorems, where we assumed $\|\nabla f_s(x) - \nabla f(x)\| \leq \mathcal{V}$ is uniformly bounded for all $x$ in the domain.

*Remark* 5.2. Such difference is negligible, because by concentration, if $f_s$ is a sample from a subgaussian stochastic estimator with variance proxy $\mathcal{V}^2$, then $\|\nabla f_s(x) - \nabla f(x)\| \leq O(\mathcal{V}\sqrt{\log(mT)})$ with overwhelming probability. As a result, this impacts our lower bounds only by a $\log(mT)$ factor. For simplicity of exposition, we only state our theorems in subgaussian stochastic estimators.

In both lower bounds, we reduce the problem to the following well-known hardness result:

**Lemma 5.3** (folklore, c.f. Lemmata 16 and 17 in [13]). *Given any $\alpha \in (0, 0.1)$, and two $T$-dimensional Gaussians $D_1 = \mathcal{N}(\mu_1, \mathbf{I})$ and $D_2 = \mathcal{N}(\mu_2, \mathbf{I})$ with $\|\mu_1 - \mu_2\| = O(\alpha)$. There exist two distributions $N_1$ and $N_2$ over $T$-dimensional vectors so that, given random samples $X_1, \ldots, X_m$ defined as below, there is no algorithm that can distinguish between the two cases with probability $\geq 2/3$:*

- CASE 1: $X_i \sim N_1$ *if* $i \in S$ *and* $X_i \sim D_1$ *if* $i \notin S$ — *where $S$ is a uniformly random subset of $[m]$ with $|S| = \alpha m$.*
- CASE 2: $X_i \sim N_2$ *if* $i \in S$ *and* $X_i \sim D_2$ *if* $i \notin S$ — *where $S$ is a uniformly random subset of $[m]$ with $|S| = \alpha m$.*

## 5.1 Lower Bound for Non-Strongly Convex Objectives

In this section, we show:

> **Theorem 5.4.** *For any $D, \mathcal{V}, \varepsilon > 0$ and $\alpha \in (0, 0.1)$, there exists a linear function $f : [-D, D] \to \mathbb{R}$ (of Lipscthiz continuity $G = \varepsilon/D$) with a subgaussian stochastic estimator with variance proxy $\mathcal{V}^2$ so that, given $m$ machines, of which $\alpha m$ are Byzantine, and $T$ samples from the stochastic estimator per machine, no algorithm can output $x$ so that $f(x) - f(x^*) < \varepsilon$ with probability $\geq 2/3$ unless*
> $$T = \Omega\left(\frac{D^2 \mathcal{V}^2}{\varepsilon^2 m} + \frac{\alpha^2 \mathcal{V}^2 D^2}{\varepsilon^2}\right), \tag{5.1}$$
> *where $x^* = \arg\min_{x \in [-D, D]} f(x)$.*

*Proof of Theorem 5.4.* The first term in (5.1) is standard for (non-Byzantine) stochastic optimization, see for instance Woodworth and Srebro [37], so it suffices to prove $T = \Omega\left(\frac{\alpha^2 \mathcal{V}^2 D^2}{\varepsilon^2}\right)$. We do so by reducing to Lemma 5.3.

Consider two linear functions $f_+(x) = \varepsilon x/D$, and $f_-(x) = -\varepsilon x/D$. The following properties are straightforward to verify:

- In the domain $[-D, D]$, $f_+(x)$ is minimized at $x = -D$ and $f_-(x)$ is minimized at $x = D$.
- For $x \in [-D, D]$, we have $f_+(x) - f(-D) < \varepsilon \iff x < 0$ and $f_-(x) - f(D) < \varepsilon \iff x > 0$.
- Define $f_s(x) = s \mathcal{V} x$. If $s \sim \mathcal{D}_+ \stackrel{\text{def}}{=} \mathcal{N}(\frac{\varepsilon}{D\mathcal{V}}, 1)$, then $f_s$ is a subgaussian stochastic estimator for $f_+(x)$ with variance proxy $\mathcal{V}^2$. Similarly, if $s \sim \mathcal{D}_- \stackrel{\text{def}}{=} \mathcal{N}(-\frac{\varepsilon}{D\mathcal{V}}, 1)$, then $f_s$ is a subgaussian stochastic estimator for $f_-(x)$ with variance proxy $\mathcal{V}^2$.

Consider any algorithm $\mathcal{A}$ for Byzantine stochastic optimization that achieves the $\varepsilon$-approximation guarantee. We claim that this gives us an algorithm $\mathcal{A}'$ for the distinguishing the two cases in Lemma 5.3 with $\|\mu_1 - \mu_2\| = \frac{2\varepsilon \sqrt{T}}{\mathcal{V} D}$ and the same $m, \alpha, T$. If so, then Lemma 5.3 directly implies

$$\frac{2\varepsilon \sqrt{T}}{\mathcal{V} D} \geq \Omega(\alpha),$$

which gives our desired lower bound.

All that is left to do is to construct the algorithm $\mathcal{A}'$. Let $\mu_1 = -\frac{\varepsilon}{D\mathcal{V}} \mathbb{1}_T$ and $\mu_2 = \frac{\varepsilon}{D\mathcal{V}} \mathbb{1}_T$ where $\mathbb{1}_T \in \mathbb{R}^T$ is the all-one vector. Our simulation proceeds as follows. Given $m$ vectors $s_1, \ldots, s_m$ where the $s_1, \ldots, s_m$ come from one of the two cases in Lemma 5.3, we give one vector to each worker machine. Then, in iteration $k \in [T]$, each machine $i$ returns a sample function $f_{s_i[k]} = s_i[k] \cdot \mathcal{V} \cdot x$. Let $S$ be as in Lemma 5.3, and let $\mathsf{good} = [m] \setminus S$. Then, by the properties explained above, for every $i \in \mathsf{good}$, in every iteration, their sample is an independent stochastic estimator for $f_+$ in Case 1 and $f_-$ in Case 2, with variance proxy $\mathcal{V}^2$, and $|\mathsf{good}| = (1-\alpha)m$. Therefore, given samples $s_1, \ldots, s_m$ from Case 1 (resp. Case 2), we can run $\mathcal{A}$ to minimize $f_+$ (resp. $f_-$).

Then, by the $\varepsilon$-approximate minimization guarantee of $\mathcal{A}$, we know that after $T$ iterations, with probability at least $\geq 2/3$, we must either return $x < 0$ and conclude that $f = f_+$, or return $x > 0$ and conclude $f = f_-$. In the former case we can conclude that the sample vector $s_1, \ldots, s_m$ are drawn from $\mathcal{N}(\mu_1, \mathbf{I})$ and $N_1$, and in the latter we conclude they are drawn from $\mathcal{N}(\mu_2, \mathbf{I})$ and $N_2$. This gives an algorithm $\mathcal{A}'$ for distinguishing these two cases in Lemma 5.3. $\square$

## 5.2 Lower Bound for Strongly Convex Objectives

In this section we show a lower bound for strongly convex objectives. A very similar construction also appeared in [39]. Formally, we show:

**Theorem 5.5.** *For any $\mathcal{V}, \sigma > 0$ and $\alpha \in (0, 0.1)$, there exists a $\sigma$-strongly convex quadratic function $f : \mathbb{R} \to \mathbb{R}$ with a subgaussian stochastic estimator of variance proxy $\mathcal{V}^2$ so that, given $m$ machines, of which $\alpha m$ are Byzantine, and $T$ samples from the stochastic estimator per machine, no algorithm can output $x$ so that $|x - x^*| < \widehat{\varepsilon}$ with probability $\geq 2/3$ unless*

$$T = \Omega\left(\frac{\mathcal{V}^2}{m\sigma^2 \widehat{\varepsilon}^2} + \frac{\alpha^2 \mathcal{V}^2}{\sigma^2 \widehat{\varepsilon}^2}\right), \tag{5.2}$$

*where $x^* = \arg\min_{x \in \mathbb{R}} f(x)$.*

**Corollary 5.6.** *Since $f(x) - f(x^*) \leq \varepsilon = \frac{\sigma \widehat{\varepsilon}^2}{2}$ implies $\|x - x^*\| \leq \widehat{\varepsilon}$ by the strong convexity of $f$, Theorem 5.5 also implies that no algorithm can output $x$ so that $f(x) - f(x^*) \leq \varepsilon$ with probability $\geq 2/3$ unless*

$$T = \Omega\left(\frac{\mathcal{V}^2}{m\sigma\varepsilon} + \frac{\alpha^2 \mathcal{V}^2}{\sigma\varepsilon}\right). \tag{5.3}$$

*Remark* 5.7. The lower bound of Yin et al. [39] uses essentially the same construction as we do in the proof of Theorem 5.5. However, in $d$ dimensions, they use a subgaussian estimator for $f$ with variance proxy $d\mathcal{V}^2$ (so $\mathbb{E}\left[\|\nabla f_s(x) - \nabla f(x)\|^2\right] \leq O(d\mathcal{V}^2)$). As a result, their lower bound appears to have an additional $d$ factor in it. Once re-normalized to have variance proxy $\mathcal{V}^2$, the hard instance in [39] yields exactly the same lower bound as our Theorem 5.5.

*Proof of Theorem 5.5.* As before, the first term in (5.2) is standard for non-Byzantine stochastic optimization (c.f. Woodworth and Srebro [37]) and so it suffices to prove $T \geq \Omega\left(\frac{\alpha^2 \mathcal{V}^2}{\sigma^2 \widehat{\varepsilon}^2}\right)$. We do so by reduction to Lemma 5.3.

Consider two functions $f_+(x) = \frac{\sigma}{2}(x - \widehat{\varepsilon})^2$ and $f_-(x) = \frac{\sigma}{2}(x + \widehat{\varepsilon})^2$. The following properties are straightforward to verify:

- $f_+(x)$ is minimized at $x = -\widehat{\varepsilon}$ and $f_-(x)$ is minimized at $x = \widehat{\varepsilon}$.

- The random function $f_s(x) = \frac{\sigma}{2}(x - s)^2 - \frac{\mathcal{V}^2}{2\sigma}$ for $s \sim D_\pm \overset{\text{def}}{=} \mathcal{N}(\pm \widehat{\varepsilon}, \mathcal{V}^2/\sigma^2)$ is a subgaussian stochastic estimator for $f_\pm$ with variance proxy $\mathcal{V}^2$.

Consider any algorithm $\mathcal{A}$ for Byzantine stochastic optimization that achieves the $\widehat{\varepsilon}$-approximation guarantee. We claim that this gives us an algorithm $\mathcal{A}'$ for the distinguishing the two cases in Lemma 5.3 where $\|\mu_1 - \mu_2\| = \frac{2\sigma\widehat{\varepsilon}}{\mathcal{V}}\sqrt{T}$ and the same $m, \alpha, T$. If so, then Lemma 5.3 directly implies

$$\frac{2\sigma\widehat{\varepsilon}}{\mathcal{V}}\sqrt{T} \geq \Omega(\alpha),$$

which gives our desired lower bound.

All that is left to do is to construct the algorithm $\mathcal{A}'$. Let $\mu_1 = -\frac{\varepsilon}{D\mathcal{V}}\mathbb{1}_T$ and $\mu_2 = \frac{\varepsilon}{D\mathcal{V}}\mathbb{1}_T$ where $\mathbb{1}_T \in \mathbb{R}^T$ is the all-one vector. Our simulation is very similar to the one in Theorem 5.4, and proceeds as follows. Given $m$ vectors $s_1, \ldots, s_m$ where the $s_1, \ldots, s_m$ come from one of the two cases in Lemma 5.3, we give one vector to each worker machine. Then, in iteration $k \in [T]$, each machine $i$ returns a sample function $f_{s_i[k]}(x) = \frac{\sigma}{2}(x - s_i[k])^2 - \frac{\mathcal{V}^2}{2\sigma}$. Let $S$ be as in Lemma 5.3, and let $\mathsf{good} = [m] \setminus S$. Then, by the properties explained above, for every $i \in \mathsf{good}$, in every iteration, their sample is an independent stochastic estimator for $f_+$ in Case 1 and $f_-$ in Case 2, with variance proxy $\mathcal{V}^2$, and $|\mathsf{good}| = (1-\alpha)m$. Therefore, given samples $s_1, \ldots, s_m$ from Case 1 (resp. Case 2), we can run $\mathcal{A}$ to minimize $f_+$ (resp. $f_-$).

Then, by the $\widehat{\varepsilon}$-approximate minimization guarantees of $\mathcal{A}$, we know that after $T$ iterations, with probability $\geq 2/3$, we must either return $x < 0$ and conclude that $f = f_+$, or return $x > 0$ and conclude that $f = f_-$. In the former case we can conclude that the samples $s_1, \ldots, s_m$ are drawn from $\mathcal{N}(\mu_1, \mathbf{I})$ and $N_1$, and in the latter we conclude they are drawn from $\mathcal{N}(\mu_2, \mathbf{I})$ and $N_2$. This gives an algorithm $\mathcal{A}'$ for distinguishing these two cases in Lemma 5.3. □